\documentclass[journal,twoside,web]{ieeecolor}
\usepackage{generic}
\usepackage{cite}
\usepackage{amsmath,amssymb,amsfonts}
\usepackage{algorithmic}
\usepackage{graphicx}
\usepackage{textcomp}

\usepackage[table]{xcolor}
\usepackage{xcolor}
\usepackage{stfloats}
\usepackage{soul}

\usepackage{booktabs}
\usepackage{subfigure}

\usepackage{etoolbox}
\usepackage{multirow}
\usepackage{balance}
\usepackage{array}
\newcolumntype{L}[1]{>{\raggedright\let\newline\\\arraybackslash\hspace{0pt}}p{#1}}
\newcolumntype{C}[1]{>{\centering\let\newline\\\arraybackslash\hspace{0pt}}p{#1}}
\newcolumntype{R}[1]{>{\raggedleft\let\newline\\\arraybackslash\hspace{0pt}}p{#1}}

\def\BibTeX{{\rm B\kern-.05em{\sc i\kern-.025em b}\kern-.08em
    T\kern-.1667em\lower.7ex\hbox{E}\kern-.125emX}}

\makeatletter
\patchcmd{\@makecaption}
  {\scshape}
  {}
  {}
  {}
\makeatother

\patchcmd{\thebibliography}{\chapter*}{\section*}{}{}

\usepackage{caption}
\begin{document}
%
\title{Automated Process Incorporating Machine Learning Segmentation and Correlation of Oral Diseases with Systemic Health}

\author{
    Gregory Yauney, 
    Aman Rana,
    Lawrence C. Wong,
    Perikumar Javia,
    Ali Muftu and
    Pratik Shah$^{\dagger}$
\thanks{Manuscript received September 25, 2018.}
\thanks{G. Yauney, A. Rana, P. Javia and Pratik Shah are with MIT Media Lab, Massachusetts Institute of Technology, Cambridge, MA, USA e-mail: \{gyauney,arana,pjavia,pratiks\}@media.mit.edu}%
\thanks{L. C. Wong and A. Muftu are with School of Dental Medicine, Tufts University, Boston, MA, USA e-mail: \{lawrence.wong, ali.muftu\}@tufts.edu}
\thanks{$^{\dagger}$Corresponding author: pratiks@media.mit.edu}
}

        

\maketitle

\begin{abstract}
Imaging fluorescent disease biomarkers in tissues and skin is a non-invasive method to screen for health conditions. We report an automated process that combines intraoral fluorescent porphyrin biomarker imaging, clinical examinations and machine learning  for correlation of systemic health conditions with periodontal disease. 1215 intraoral fluorescent images, from 284 consenting adults aged 18-90, were analyzed using a machine learning classifier that can segment periodontal inflammation. The classifier achieved an AUC of 0.677 with precision and recall of 0.271 and 0.429, respectively, indicating a learned association between disease signatures in collected images. Periodontal diseases were more prevalent among males (\textit{p}=0.0012) and older subjects (\textit{p}=0.0224) in the screened population. Physicians independently examined the collected images, assigning localized modified gingival indices (MGIs). MGIs and periodontal disease were then cross-correlated with responses to a medical history questionnaire, blood pressure and body mass index measurements, and optic nerve, tympanic membrane, neurological, and cardiac rhythm imaging examinations. Gingivitis and early periodontal disease were associated with subjects diagnosed with optic nerve abnormalities (\textit{p}$<$0.0001) in their retinal scans. We also report significant co-occurrences of periodontal disease in subjects reporting swollen joints (\textit{p}=0.0422) and a family history of eye disease (\textit{p}=0.0337). These results indicate cross-correlation of poor periodontal health with systemic health outcomes and stress the importance of oral health screenings at the primary care level. Our screening process and analysis method, using images and machine learning, can be generalized for automated diagnoses and systemic health screenings for other diseases.
\end{abstract}


%

\section{Introduction}

Biomarkers provide a fast, accurate, and non-invasive way to diagnose several diseases and can be used for prognostic screening along with clinical response monitoring. Gingivitis is the inflammation of gingiva around the tooth, making the gums sensitive and likely to bleed. Gingivitis can progress and lead to periodontitis, with severe inflammation and infections in the surrounding structures of the teeth. Periodontal disease is a major cause of tooth loss in adults, due to soft tissue inflammation and bone loss. Periodontal disease has been found to be an important indicator of oral-systemic health \cite{kane2017effects} and has been linked to cardiovascular disease, osteoporosis, and diabetes \cite{ramseier2009identification,giannobile2011translational}. Clinical assessment of periodontal disease is usually done using visual examinations and probing. Standard diagnostic practices of alveolar bone height and clinical attachment levels, while helpful, do not account for patient-to-patient variation or identify disease progression risk \cite{ebersole2015targeted,ramseier2009identification}.

Biomarkers indicative of periodontal and gingival disease can be detected from crevicular fluid using biochemical assays, but doing so is impractical in many clinical settings. Gingivitis results in an increased blood flow around the inflamed gingiva, leading in turn to increased red fluorescence (650 nm) from porphyrin in the surrounding vasculature. Gums when illuminated with a blue light (405-450 nm), fluoresce in the presence of porphyrin from hemoglobin in the blood due to inflammation and microbial biofilms \cite{rechmann2016performance}. Trained dental experts, without the aid of porphyrin fluorescence, can also discern periodontal diseases and gingivitis in whitelight images. Computer vision, machine learning, and deep neural networks can perform automated and accurate diagnoses of several diseases from images \cite{esteva2017dermatologist}. We have previously described an automated system that performs pixel-wise segmentation of the inflamed gingiva to detect gingivitis and periodontal disease using fluorescent images acquired by an FDA-approved intraoral camera \cite{rana2017automated}.

In this study, we present a new medical imaging and informatics based process for demonstrating the generalizability of automated oral health screenings and cross correlations of oral-systemic health (Figure 1). Intraoral fluorescent images were collected from 284 consenting adults and analyzed for periodontal diseases using our previously described machine learning classifier. Segmentation results from the classifier were compared with localized labels provided by dentists for the same images. We then analyzed co-occurrence rates between subjects' MGIs, a measure of periodontal health provided by expert dentists, and three sources of screenings: 1) a self-reported medical history questionnaire, 2) a group of routine health screenings: blood pressure (BP) and body mass index (BMI) measurements, and 3) a group of technology-enabled screenings (TES): single-lead echocardiogram (ECG) arrhythmias, tympanic membrane disorders, blood oxygenation levels, optic nerve disorders, and neurological fitness exams conducted using FDA approved devices. Higher MGIs were significantly correlated with males, older age, swollen joints, and a family history of eye disease. Gingivitis was significantly correlated with optic nerve exam abnormalities. Our automated process and results thus indicate that periodontal health is an important aspect of systemic and overall subject health. 

\begin{figure}[t!]
\begin{center}
\includegraphics[width=\linewidth]{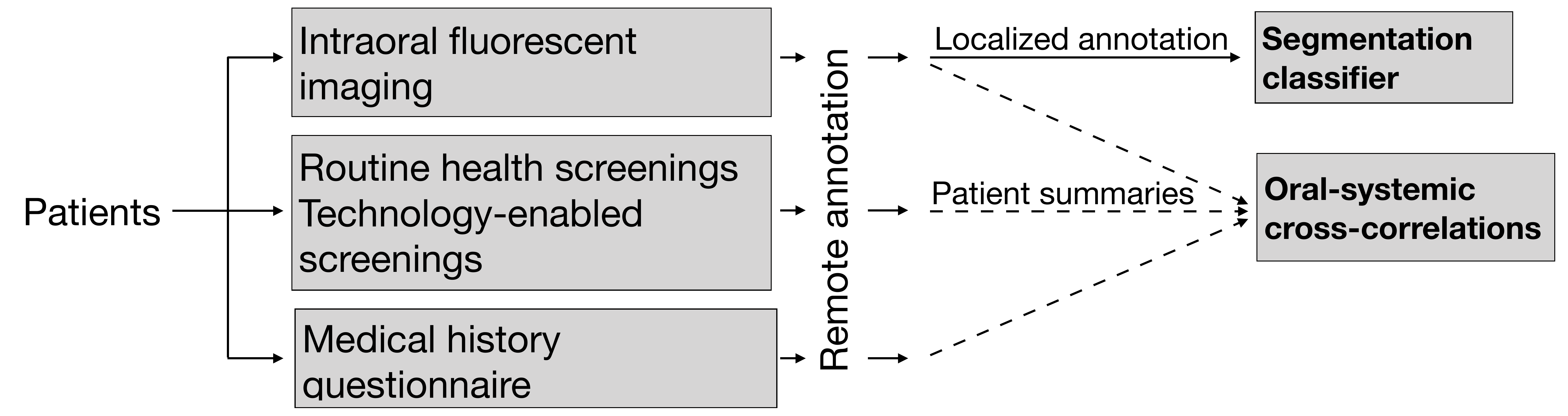}
\caption{The process starts with non-invasive imaging and health screening in an out-patient setting. Expert annotations of localized and patient-level condition signatures were used to evaluate a segmentation classifier and oral-systemic cross-correlations. Routine health screenings include blood pressure and body mass index measurements, and technology-enabled screenings consist of imaging examinations of blood oxygenation levels, optic nerve, tympanic membrane, coordination, and gait.}
\label{figure3}
\end{center}
\vskip -0.2in
\end{figure}

\section{Related work}

\subsection{Convolutional neural networks}
Deep neural networks allow for representation learning and complex feature extraction. Convolutional neural networks (CNNs) require fewer total parameters because matrix multiplication is replaced by convolution, with each neuron sharing weights. The convolutional layers are followed by non-linear functions which are in turn generally followed by pooling layers. The initial layers extract spatial features, and pooling provides translational invariance, making CNNs robust for visual classification tasks \cite{krizhevsky2012imagenet}.

\subsection{Medical image segmentation and disease diagnosis process}

CNNs are widely used for automatic detection and segmentation of various diseases in medical images. CNNs have recently been used to segment lesions from computed tomography images and magnetic resonance imaging of the brain and liver \cite{gao2018segmentation}. Recent work on medical image segmentation includes using a CNN for classification of cancerous oral cavity regions and classification in quantitative light-induced images \cite{aubreville2017automatic,imangaliyev2017classification}. CNNs have also been used explicitly for gingival health tasks with intraoral whitelight and fluorescence images, such as full-image segmentation of gingivitis and patch-based segmentation of plaque and periodontal diseases \cite{rana2017automated,yauney2017convolutional}.

Medical imaging, especially low-cost imaging provided by mobile phones and specialized imaging devices, provides non-invasive measures of patient health and has been used for applications of remote diagnosis in addition to the segmentation work described above \cite{granot2008new,angelino2017clinical}. Past work on remote and automated diagnosis of medical images did not afford a holistic consideration of the systemic health of patients whose images were analyzed \cite{litjens2017survey}. Automated processes integrating imaging and remote annotation informatics for diagnosis and systemic health analysis have recently been described by us \cite{Shahe018774}. These processes provide opportunities for gathering the data required to build accurate machine learning models for disease segmentation while also providing the ability to link systemic health of the patients to images.

\subsection{Periodontal health epidemiology}

 Periodontal disease is correlated with coronary heart disease and peripheral vascular disease in American adults \cite{destefano1993dental,mendez1998association}. The former was especially more likely in young adult males than in other age and gender cohorts \cite{destefano1993dental}. Significant links between periodontal disease and diabetes, both types I and II, were found in multiple studies of thousands of Pima Indians of the Gila River Indian Community in Arizona, US, and in Danish men \cite{loe1993periodontal}. Periodontal disease was correlated with obesity in Japanese adults irrespective of age and, in a sample of over ten thousand American adults, in younger American adults. In the same study, low BMI was significantly associated with decreased prevalence of periodontal disease among young American adults \cite{al2003obesity}. Separate studies of thousands of American adults found that recent tooth loss was significantly correlated with primary open-angle glaucoma and that periodontal disease was significantly correlated with macular degeneration in adults under the age of 60 \cite{pasquale2016prospective,wagley2015periodontal}. In various populations in India, periodontal disease has been found to be highly prevalent and significantly correlated with older age, coronary artery disease, tobacco use, smoking, and lower socioeconomic status \cite{shewale2016prevalence,shaju2011prevalence}. However, all the above studies relied on clinical examinations by human experts and none used intraoral images, TES, or machine learning for diagnosing periodontal diseases to detect novel oral-systemic cross correlations.

\section{Methods}

\subsection{Data acquisition}

Data from 284 consenting adults aged 18-90 in Maharashtra, India, was used for this study. The Mahatma Gandhi Vidyamandir Karmaveer Bhausaheb Hiray Dental College \& Hospital institutional ethics committee reviewed and approved protocol MGVKBHDC/15-16/571 for clinical data collection. De-identified data was transferred and analyzed at the Massachusetts Institute of Technology (MIT) in Cambridge, MA, according to MIT Committee on the Use of Humans as Experimental Subjects approval for protocol 1512338971. SOPROCARE (SOPRO Acteon Imaging, France) was used for intraoral imaging using white light and 405nm light-emitting diodes. Optic nerve, ECG, tympanic membrane, oxygen saturation, gait and coordination exams and BMI and blood pressure measurements recorded with FDA-approved devices were used for oral-systemic cross correlations \cite{Shahe018774}.

\subsection{Data preprocessing and clinical examinations}

De-identified data assigned to unique subject identifiers was split into separate pools consisting of optic nerve, tympanic membrane, periodontal, and neurological images for all study participants. BMI and blood pressure are routinely measured by most primary care providers and have been collectively annotated as routine health screenings throughout this study. Other imaging and smartphone-based tests have been designated TES methods. Routine health screenings and responses to medical questionnaires were grouped together for computational analyses. For BMI, numbers less than 19 were labeled low, between 19 and 25 were characterized as normal, and 25 and above were considered high (Panel, 1998). For blood pressure, systolic pressure below 90 mmHg or diastolic pressure below 60 mmHg was considered low, systolic pressure between 90 and 140 mmHg and diastolic pressure between 60 and 90 mmHg was labeled normal, and systolic pressure above 140 mmHg or diastolic pressure above 90 mmHg was labeled high \cite{pickering2005recommendations}. Blood oxygen levels of 90\% or less were annotated low. 

Optic nerve, tympanic membrane, gait, and coordination images captured in TES were categorized by patient ID and TES examination and displayed directly to expert physicians via a web-based examination portal \cite{rana2017automated, Shahe018774}. Annotators were able to mark specific clinical features that were present in each video, and a condition was kept if it was noted by a majority of experts \cite{Shahe018774}. The outputs from the AliveCor mobile app were readily used as annotations for ECG tests because they were labeled `Normal' or `Possible atrial fibrillation' \cite{nitzan2014pulse}.

\subsection{Periodontal examinations using images}

Using the same web-based examination portal, dental experts examined de-identified intraoral fluorescence images and annotated periodontal disease on the gingival margin and left and right papillae. The experts also assigned each image a modified gingival index (MGI) ranging from 0 (healthy gingiva) to 5 (severe periodontal disease) \cite{lobene1986modified}. A majority MGI was calculated for each subject. For subjects with no clear majority, the greater tied MGI was taken so as to not understate the prevalence of periodontal disease.

\subsection{Disease segmentation}

In a previously described study we used 405 (from a total 1215) intraoral images for training and validation of a segmentation classifier \cite{rana2017automated}. The model accepts an RGB image of size $640\times480\times3$ and outputs a binary mask indicating the locations of periodontal disease. In this study the trained model was used to segment periodontal disease in additional 810 images. Ground truth segmentation masks were prepared by spatially bounding the locations of all expert annotations for all 810 images. The bounded region was thresholded based on the augmented color signature of periodontal disease to capture finer signatures of inflammation. We evaluated classifier performance by calculating standard measures like true and false positive rates, precision, recall, and mean intersection over union (IOU).

\begin{table*}[b!]
 \resizebox{\linewidth}{!}{%
 \begin{tabular}{l@{\hskip 1cm}cccc@{\hskip 1cm}cccc@{\hskip 1cm}cccc@{\hskip 1cm}cccc@{\hskip 1cm}ccc}
\toprule
& \multicolumn{3}{c}{Adolescent (18-19)} & & \multicolumn{3}{c}{Young adult (20-39)} & & \multicolumn{3}{c}{Middle age (40-64)} & & \multicolumn{3}{c}{Old age (65-90)} & & \multicolumn{3}{c}{\textbf{All ages}}\\
MGI & Female & Male & Total & & Female & Male & Total & & Female & Male & Total & &Female & Male & Total & & Female & Male & \textbf{Total} \\
\midrule
0 & 1 & 0 & 1 & & 0 & 0 & 0 & & 0 & 1 & 1 & & 0 & 0 & 0 & & 1 & 1 & \textbf{2}\\
1 & 9 & 4 & 13 & & 10 & 5 & 15 & & 3 & 5 & 8 & & 0 & 3 & 3 & & 22 & 17 & \textbf{39}\\
2 & 20 & 8 & 28 & & 17 & 23 & 40 & & 19 & 21 & 40 & & 2 & 10 & 12 & & 58 & 62 & \textbf{120}\\
3 & 3 & 7 & 10 & & 10 & 21 & 31 & & 9 & 24 & 33 & & 3 & 15 & 18 & & 25 & 67 & \textbf{92}\\
4 & 0 & 0 & 0 & & 0 & 5 & 5 & & 8 & 8 & 16 & & 2 & 7 & 9 & & 10 & 20 & \textbf{30}\\
5 & 0 & 0 & 0 & & 0 & 0 & 0 & & 1 & 0 & 1 & & 0 & 0 & 0 & & 1 & 0 & \textbf{1}\\
\bottomrule
 \end{tabular}
 }
 \caption{The distribution of patient-level modified gingival indices (MGIs) split by gender and age cohort. The majority of subjects (74.6\%) had MGIs of 2 or 3, indicating prevalent gingival diseases.}
\end{table*}

\subsection{Cross-correlations and statistical methods}

Since periodontal disease segmentations on images and localized annotations have not been clinically used to assign MGIs and are too fine-grained, we used expert-assigned MGIs for correlation of imaging results with systemic health outcomes. We investigated correlations between periodontal health MGIs provided by expert dentists and gender and age across the dataset. Correlations between periodontal health MGIs and other conditions identified by 1) the medical history questionnaire, 2) routine health screenings, and 3) TES tests were also analyzed. We first calculated the numbers of patients with all possible co-occurrences of each MGI and condition identified by the three groups of screenings. We then compared incidence rates between each co-occurrence and between genders and age cohorts. Fisher's exact test was used to establish statistical significance when comparing between two populations with unequal numbers of subjects and possibly unequal means and variances, and we accept as significant all comparisons with a p-value less than 0.05. For example, to find that subjects with an MGI of 4 are more likely to also have reported swollen joints than subjects with other MGIs, we calculate two ratios: the number of subjects who reported swollen joints and have an MGI of 4 (14) over the number of subjects with an MGI of 4 (30), and the number of subjects who reported swollen joints and have an MGI that is not 4 (56) over the number of subjects with an MGI that is not 4 (254). Comparing the two ratios with Fisher's exact test yields a \textit{p}-value of 0.0422, and we conclude the correlation is significant. We separately investigated the extent to which automated MGI prediction could provide an end-to-end pipeline for MGI and correlated systemic health prediction.

\section{Results} 

\subsection{Segmentation of periodontal disease in subjects by machine learning classifier}

The classifier was tested on 810 images, producing a true positive rate of 0.429 and a false positive rate of 0.075 when individual predictions for all pixels in each image are considered. The area under the resulting receiver operating characteristic curve was 0.677, interpretable as a 67.7\% chance that a pixel labeled with periodontal disease would be more likely to be predicted as periodontal disease than a pixel labeled as healthy. Precision was 0.271. Figure 2 shows these values in the form of a receiver operating characteristic (ROC) curve and a precision-recall curve, both in context with baseline random chance, which the classifier outperforms. Mean intersection over union (IOU) was 0.1710 $\pm$ 0.1544 when averaged across images. The classifier robustly segments numerous configurations of gingival and periodontal disease (Figure 3, columns i-iv). When segmenting an image of healthy gingiva, the classifier predicts a commensurately smaller disease extent though it does produce some false positives (Figure 3, column v). On rare occasions, the classifier understates the extent of disease (Figure 3, column vi).

\begin{figure}[t!]
\begin{center}
\subfigure[]{%
       \includegraphics[width=.7\columnwidth]{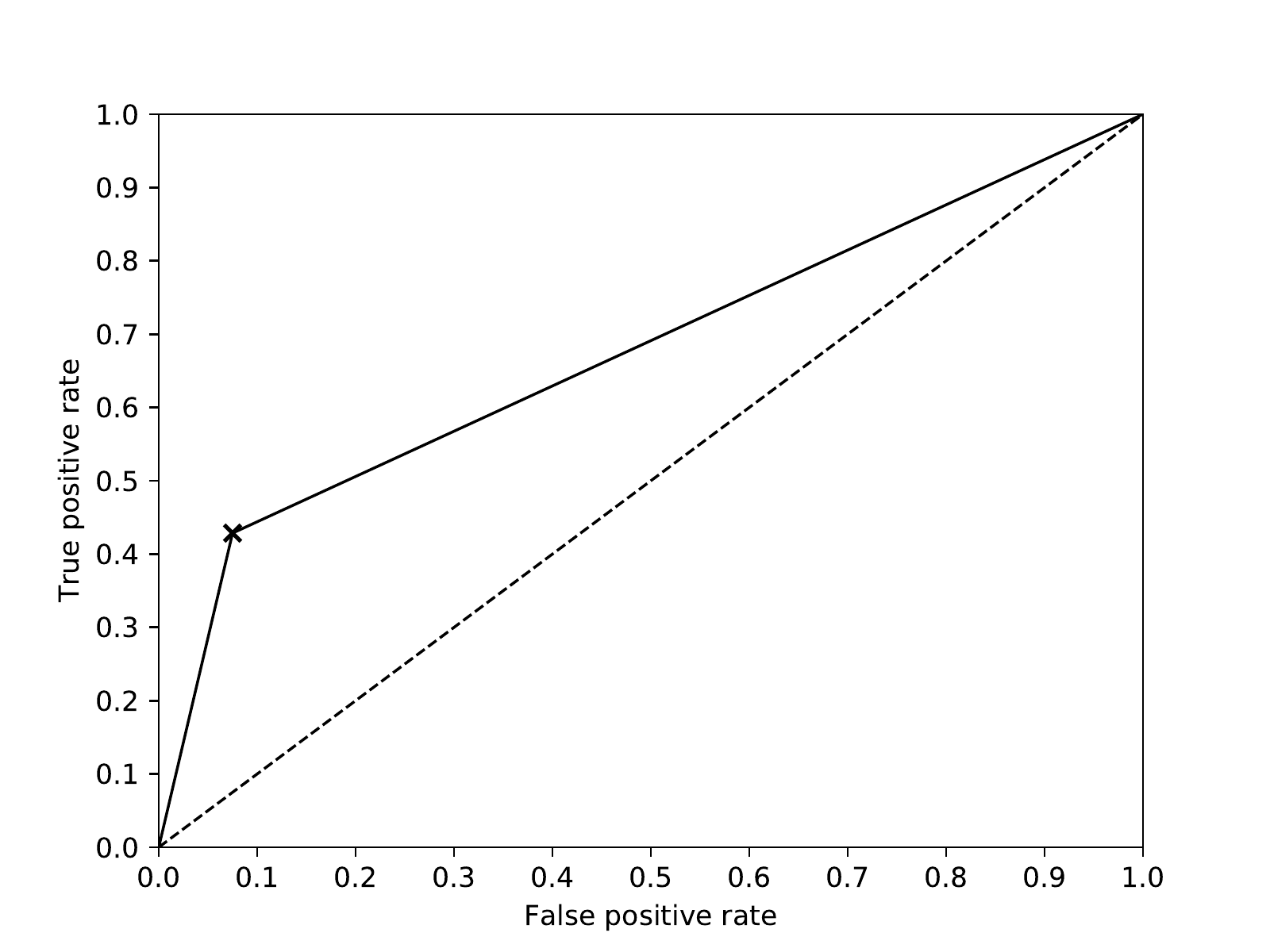}}
    \label{2a}
 \subfigure[]{%
       \includegraphics[width=.7\columnwidth]{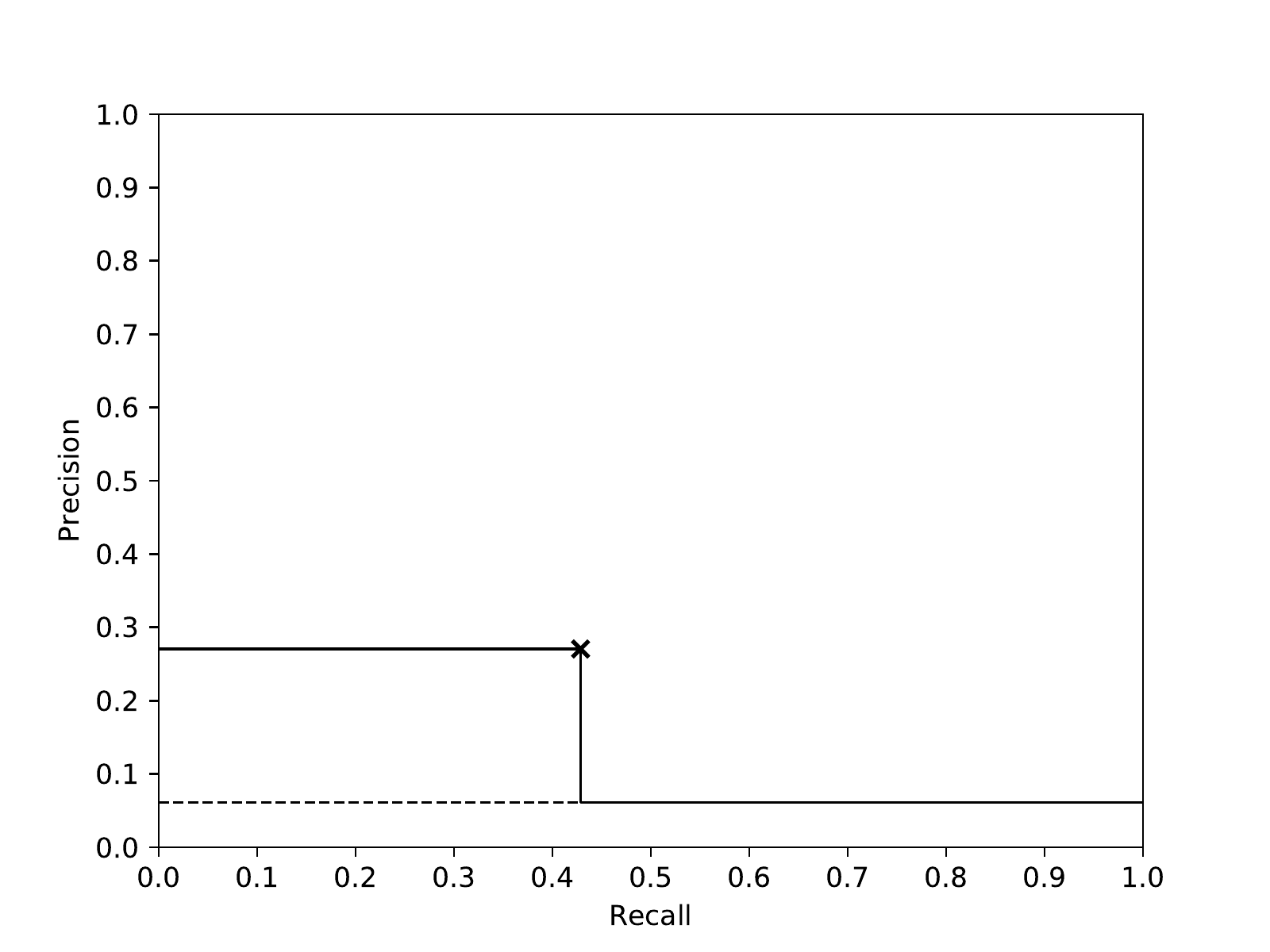}}
    \label{2b}
\caption{The trained segmentation classifier, shown as a solid line, outperforms random chance, shown as a dashed line, on 810  intraoral images. The classifier is marked in both graphs. (a) Receiver operating characteristic (ROC) curve. The marked point occurs at a false positive rate of 0.075 and a true positive rate of 0.429. Area under the ROC curve is 0.677. (b) Precision-recall curve. The marked point occurs at a recall of 0.429 and a precision of 0.271.}
\label{figure3}
\end{center}
\vskip 0.3in
\end{figure}

\renewcommand{\thesubfigure}{(\roman{subfigure})}
\begin{figure}[t!]
\begin{flushright}
\subfigure{\makebox[0pt][r]{\makebox[30pt]{\raisebox{15pt}{(A)}}}\includegraphics[width=.14\columnwidth]{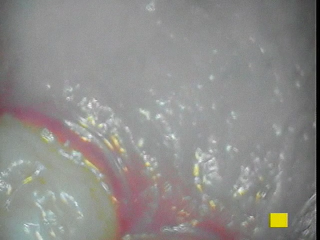}}
\subfigure{\includegraphics[width=.14\columnwidth]{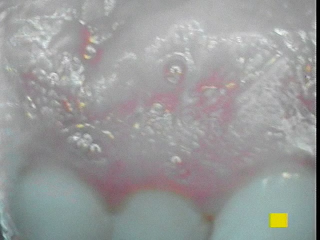}}
\subfigure{\includegraphics[width=.14\columnwidth]{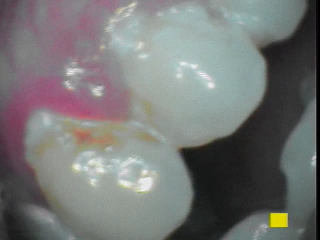}}
\subfigure{\includegraphics[width=.14\columnwidth]{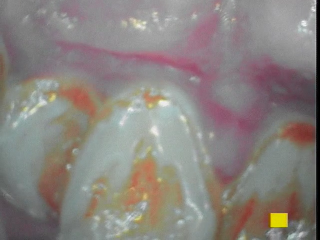}}
\subfigure{\includegraphics[width=.14\columnwidth]{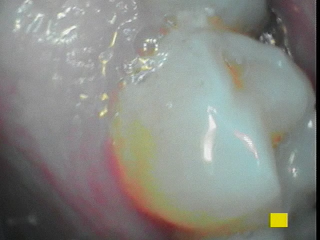}}
\subfigure{\includegraphics[width=.14\columnwidth]{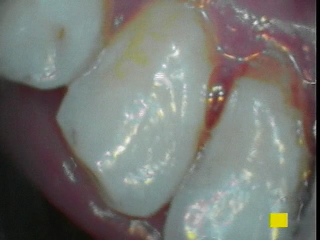}}\\[-3mm]
\subfigure{\makebox[0pt][r]{\makebox[30pt]{\raisebox{15pt}{(B)}}}\includegraphics[width=.14\columnwidth]{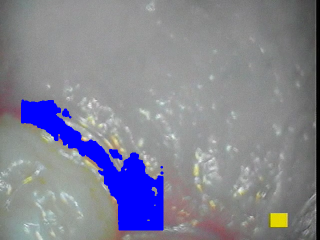}}
\subfigure{\includegraphics[width=.14\columnwidth]{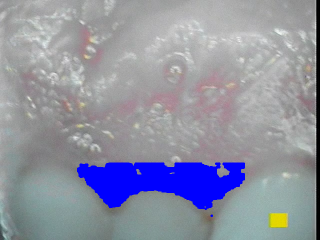}}
\subfigure{\includegraphics[width=.14\columnwidth]{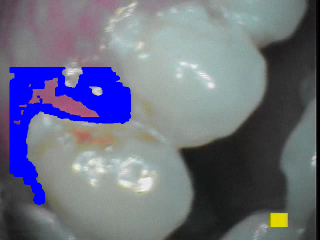}}
\subfigure{\includegraphics[width=.14\columnwidth]{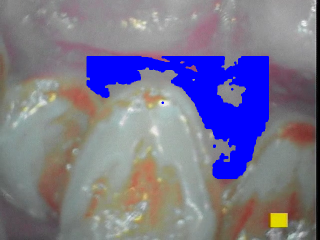}}
\subfigure{\includegraphics[width=.14\columnwidth]{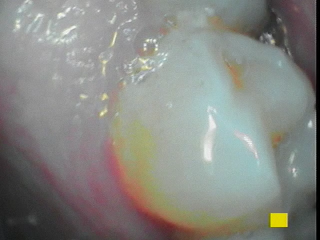}}
\subfigure{\includegraphics[width=.14\columnwidth]{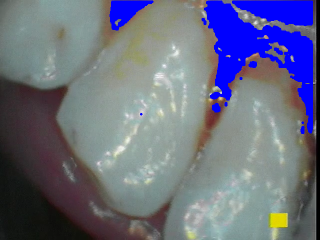}}\\[-3mm]
\subfigure{\makebox[0pt][r]{\makebox[30pt]{\raisebox{15pt}{(C)}}}\includegraphics[width=.14\columnwidth]{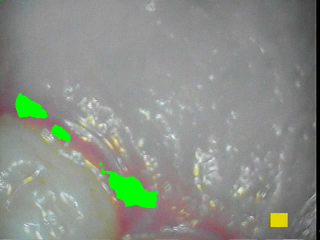}}
\subfigure{\includegraphics[width=.14\columnwidth]{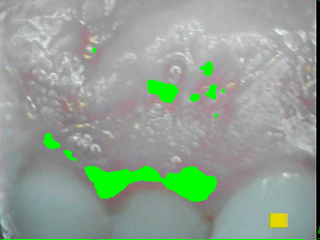}}
\subfigure{\includegraphics[width=.14\columnwidth]{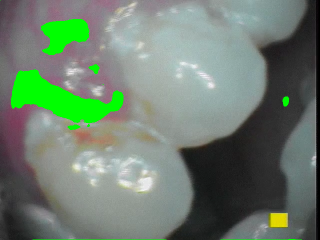}}
\subfigure{\includegraphics[width=.14\columnwidth]{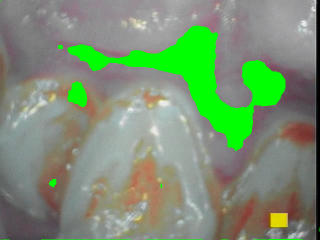}}
\subfigure{\includegraphics[width=.14\columnwidth]{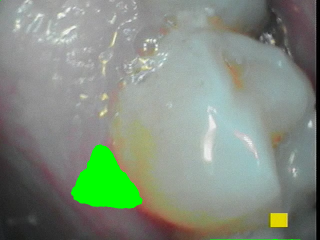}}
\subfigure{\includegraphics[width=.14\columnwidth]{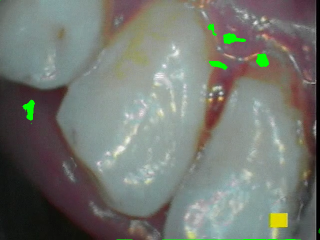}}\\[-3mm]
\setcounter{subfigure}{0}
\subfigure[]{\makebox[0pt][r]{\makebox[30pt]{\raisebox{15pt}{(D)}}}\includegraphics[width=.14\columnwidth]{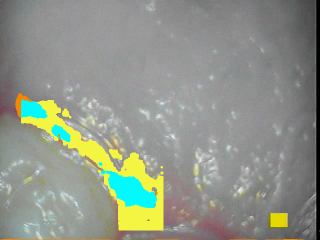}}
\subfigure[]{\includegraphics[width=.14\columnwidth]{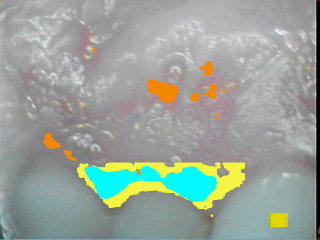}}
\subfigure[]{\includegraphics[width=.14\columnwidth]{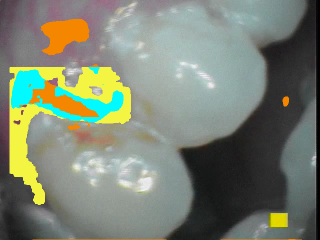}}
\subfigure[]{\includegraphics[width=.14\columnwidth]{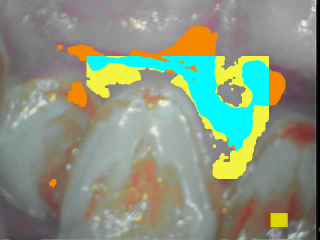}}
\subfigure[]{\includegraphics[width=.14\columnwidth]{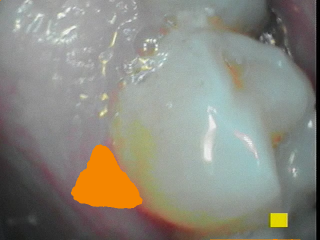}}
\subfigure[]{\includegraphics[width=.14\columnwidth]{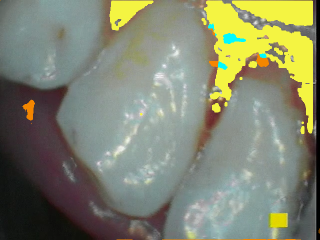}}\\[-1mm]
\end{flushright}
\begin{center}
\subfigure{\includegraphics[width=0.7\columnwidth]{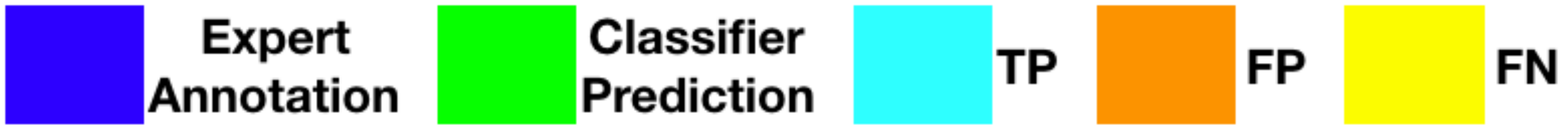}}\\[-3mm]
\caption{Representative segmentation inputs and results. Row (A): intraoral fluorescence images captured during screening; row (B): corresponding ground truth signatures of periodontal disease constructed from localized expert annotation; row (C): corresponding predictions from the segmentation classifier; row (D): visualization of localized prediction errors. Each column contains an image from a different subject.}
\label{figure3}
\end{center}
\end{figure}
\renewcommand{\thesubfigure}{(\alph{subfigure})}

\subsection{Periodontal health of subjects using MGI scores}

Table 1 shows summary MGI scores for all 284 subjects, aggregated across all images from each subject. 42.3\% of subjects had an aggregated MGI of 2 and another 32.4\% had an MGI of 3, indicating a moderate presence of gingival diseases in the majority of subjects. Two subjects were deemed to be in good gingival health with an MGI of 0. Females were more likely than males to have an MGI of 2 (\textit{p}=0.0389), and males were more likely than females to have an MGI of 3 (\textit{p}=0.0012). This indicates that males in our dataset tended to have higher MGIs and periodontal diseases than females. Higher MGIs were correlated with middle-aged (\textit{p}=0.0013, \textit{p}=0.0213) and old-aged (\textit{p}=0.0224, \textit{p}=0.0004) cohorts.

\subsection{ Periodontal health correlations: medical history questionnaire}

Table 2 shows the number of subjects with a particular MGI score and who answered yes to each question on the medical history questionnaire. Subjects with an MGI of 4 were more likely to report swollen joints (\textit{p}=0.0422) and a family history of eye disease (\textit{p}=0.0337).

 \begin{table*}[ht!]
 \resizebox{\linewidth}{!}{%
 \begin{tabular}{ll@{\hskip 0.3cm}cccccccccccccccccccccccccc}
\toprule
MGI & \rotatebox{70}{No. patients} & \rotatebox{70}{Glasses} & \rotatebox{70}{Dental}& \rotatebox{70}{Swollen joints} & \rotatebox{70}{Hearing} & \rotatebox{70}{FH diabetes} & \rotatebox{70}{FH high BP} & \rotatebox{70}{Tobacco} & \rotatebox{70}{Difficulty walking} & \rotatebox{70}{High BP} & \rotatebox{70}{Diabetes} & \rotatebox{70}{High BP Rx} & \rotatebox{70}{Asthma} & \rotatebox{70}{Smoking} & \rotatebox{70}{FH cardiac} & \rotatebox{70}{Cardiac Rx} & \rotatebox{70}{Cardiovascular} & \rotatebox{70}{Low BP} & \rotatebox{70}{FH stroke} & \rotatebox{70}{FH eye disease} & \rotatebox{70}{Heart attack} & \rotatebox{70}{Coronary bypass} & \rotatebox{70}{Drinking} & \rotatebox{70}{Eye treatment} & \rotatebox{70}{Memory loss} & \rotatebox{70}{Ear treatment} & \rotatebox{70}{FH ear disease}\\
\midrule
0 & 2 & 1 (50.0) & 0 (0) & 0 (0) & 0 (0) & 1 (50.0)	& 0 (0) & 1 (50.0) & 0 (0) & 0 (0) & 1 (50.0) & 0 (0) & 0 (0) & 0 (0) & 0 (0) & 0 (0) & 0 (0) & 0 (0) & 0 (0) & 0 (0) & 0 (0) & 0 (0) & 0 (0) & 0 (0) & 0 (0) & 0 (0) & 0 (0)\\
1 & 39 & 23 (59.0) & 11 (28.2) & 4 (10.3) & 7 (18.0) & 13 (33.3) & 7 (18.0) & 1 (2.6) & 3 (7.7) & 1 (2.6) & 1 (2.3) & 1 (2.3) & 3 (7.7) & 1 (2.3) & 3 (7.7) & 0 (0) & 0 (0) & 0 (0) & 1 (2.6) & 0 (0) & 0 (0) & 1 (2.6) & 0 (0) & 0 (0) & 0 (0) & 1 (2.6) & 0 (0)\\
2 & 120 & 59 (49.2) & 27 (22.5) & 23 (19.2) & 17 (14.2) & 30 (25.0) & 22 (18.3) & 7 (5.8) & 8 (6.7) & 6 (5.0) & 8 (6.7) & 5 (4.2) & 5 (4.2) & 3 (2.5) & 2 (1.7) & 2 (1.7) & 1 (0.8) & 1 (0.8) & 2 (1.7) & 0 (0) & 0 (0) & 0 (0) & 1 (0.8) & 0 (0) & 0 (0) & 1 (0.8) & 0 (0)\\
3 & 92 & 43 (46.7) & 25 (27.2) & 29 (31.5) & 22 (23.9) & 12 (13.0) & 12 (13.0) & 11 (12.0) & 12 (13.0) & 8 (8.7) & 7 (7.6) & 6 (6.5) & 2 (2.2) & 2 (2.2) & 3 (3.3) &3 (3.3) & 1 (1.1) & 2 (2.2) & 1 (1.1) & 1 (1.1) & 2 (2.2) & 0 (0)& 1 (1.1) & 0 (0) & 0 (0) & 0 (0) & 0 (0)\\
4 & 30 & 15 (50.0) & 8 (26.7) & \textbf{14 (46.7)*} & 11 (36.7) & 5 (16.7) & 2 (6.7) & 3 (10.0) & 5 (16.7) & 2 (6.7) & 4 (13.3) & 1 (3.3) & 1 (3.3) & 3 (10.0) & 	0 (0) & 1 (3.3) & 0 (0) & 0 (0) & 0 (0) & \textbf{2 (6.7)*} & 0 (0) & 0 (0) & 0 (0) & 1 (3.3) & 0 (0) & 0 (0) & 1 (3.3)\\
5 & 1 & 0 (0) & 0 (0) & 0 (0) & 0 (0) & 0 (0) & 0 (0) & 0 (0) & 0 (0) & 0 (0) & 0 (0) & 0 (0)	& 0 (0) & 0 (0) & 0 (0) & 0 (0) & 0 (0) & 0 (0) & 0 (0) & 0 (0) & 0 (0) & 0 (0) & 0 (0) & 0 (0) & 0 (0) & 0 (0) & 0 (0)\\
\bottomrule
 \end{tabular}
 }
 \caption{Numbers and percentages (in parentheses) of subjects with each modified gingival index (MGI) who responded yes to each question on a medical history questionnaire. $^* p < 0.05$, shown in bold: subjects with the row's MGI are more likely to have responded yes to the column's question than subjects with other MGIs. BP: blood pressure; FH: family history; Rx: treatment.
}
\vskip-.1in
\end{table*}

\begin{table*}[t!]
 \resizebox{\linewidth}{!}{%
 \begin{tabular}{lc@{\hskip 1cm}ccccc@{\hskip 1cm}ccccc}
\toprule
& & \multicolumn{4}{c}{Routine health screenings} & & \multicolumn{5}{c}{Technology-enabled screenings}\\
MGI & No. patients & High BP & Low BP& High BMI & Low BMI & & Low O$_2$ & Retinal & TM & Finger-nose & Gait\\
\midrule
0 & 2 & 1 (50.0) & 0 (0) & 2 (100) & 0 (0) & & 0 (0) & 0 (0) & 0 (0) & 0 (0) & 0 (0)\\
1 & 39 & 4 (10.3) & 1 (2.6) & 18 (46.2) & 5 (12.8) & &
1 (2.6) & \textbf{5 (12.8)*} & 3 (7.7) & 0 (0) & 0 (0)\\
2 & 120 & 24 (20.0) & 0 (0) & 55 (45.8) & 19 (15.8) & &
6 (5.0)	 & 0 (0) & 8 (6.7) & 0 (0) & 0 (0)\\
3 & 92 & 14 (15.2) & 2 (2.2) & 11 (12.0) & 18 (19.6) & &
4 (4.4) & 0 (0) & 10 (10.9) & 2 (2.2) & 1 (1.1)\\
4 & 30 & 9 (30.0) & 0 (0) & 11 (36.7) & 7 (23.3) & &
1 (3.3) & 0 (0) & 2 (6.7) & 0 (0) & 1 (3.3)\\
5 & 1 & 0 (0) & 0 (0) & 0 (0) & 0 (0) & &
0 (0) & 0 (0) & 0 (0) & 0 (0) & 0 (0)\\
\bottomrule
 \end{tabular}
 }
 \caption{Numbers and percentages (in parentheses) of subjects with each modified gingival index (MGI) who were reported as abnormal in each routine health screening and technology-enabled screening. $* p < 0.05$, shown in bold: subjects with the row's MGI are more likely to have the column's condition than subjects with other MGIs. BP: blood pressure; BMI: body mass index; O$_2$: blood oxygen level; TM: tympanic membrane.}
 \vskip-.2in
 \end{table*}

Supplementary Table I shows the number of subjects of each gender with a particular MGI and who answered yes to each question on the medical history questionnaire. Males and females in our study both showed prevalence of periodontal diseases. Among individual genders, females with an MGI of 4 were more likely than females with other MGIs to report swollen joints (\textit{p}=0.0195), difficulty hearing (\textit{p}=0.0245), and difficulty walking (\textit{p}=0.0193). Males with an MGI of 4 were more likely than males with other MGIs to report a family history of eye diseases (\textit{p}=0.0163).

Supplementary Table II shows the number of subjects of each age cohort with a particular MGI and who answered yes to questions on the medical history questionnaire. Among middle-aged subjects, those with an MGI of 1 were more likely to report asthma MGIs (\textit{p}=0.0475). Young adults with an MGI of 4 were more likely to report a family history of eye diseases (\textit{p}=0.0049).

\subsection{ Periodontal health correlations: routine health screenings}

Table 3 shows the number of subjects with a particular MGI and who were found to have tested positive in each of the routine health screenings consisting of blood pressure and BMI measurements. Supplementary Table III shows the number of subjects of each gender with a particular MGI and who were found to have tested positive in each of the routine health screenings. Supplementary Table IV shows the number of subjects in each age cohort with a particular MGI and who were found to have tested positive in each of the routine health screenings. MGIs were not found to be significantly correlated with routine health screening outcomes in this dataset.

\subsection{Periodontal health correlations: technology-enabled screenings}

Table 3 shows the number of subjects with each MGI score who were found to have an abnormality in each of the TES tests. Subjects with an MGI of 1 were significantly more likely to have an optic nerve exam abnormality than subjects with other MGIs (\textit{p}$<$0.0001). Supplementary Table III shows the number of subjects of each gender with a particular MGI and who were found to have tested positive in each TES. Males with an MGI of 1 were more likely than males with other MGIs to also have an optic nerve exam abnormality (\textit{p}=0.0002). Supplementary Table IV shows the number of subjects in each age cohort with a particular MGI and who were found to have tested positive in each TES. Old-aged subjects with an MGI of 1 were more likely than old-aged subjects with other MGIs to have an optic nerve exam abnormality (\textit{p}=0.0002).

\section{Discussion}

\subsection{Generalization of the classifier on new dataset}

Our previously published results reported an AUC of 0.746 for segmentation of periodontal diseases in intraoral images \cite{rana2017automated}. The 0.677 AUC we report for this dataset is consistent with expected generalization error and indicates a good performance. Precision and recall experienced similar modest decreases, from 0.347 and 0.621 to .271 and .429, respectively \cite{rana2017automated}. Mean IOU for the classifier's previously reported validation dataset was 0.1824 $\pm$ 0.1547 when averaged across images, only slightly higher than the mean IOU of 0.1710 $\pm$ 0.1544 we report for this dataset. The difference in IOU is not significant when compared with a two-sample student's \textit{t}-test that accounts for different numbers of samples and possibly unequal variances (\textit{p}=0.4099). The classifier accurately segments many instances of periodontal disease (Figure 3, columns i-iv). When the classifier does produce false positives (Figure 3, column v), it is likely due to the high prevalence of periodontal disease in the training dataset. Specular reflection and possible disease configurations may contribute to false negatives (Figure 3, column vi). Such errors may be reduced by training the classifier with an increased number of images that capture more varied instances of periodontal disease and healthy patients. 
The ROC and precision-recall curves have few constituent points because the classifier is extremely confident in all of its predictions. This was also reported in the previous study with this classifier \cite{rana2017automated}. Regularization may help the classifier temper its prediction confidence, leading to increased generalization performance. Overall, our results further validate our previously published model \cite{rana2017automated} and demonstrate that meaningful periodontal health diagnoses can be annotated remotely and subsequently associated with fluorescent image regions.

\subsection{Oral-systemic health cross correlations}

our sample of 284 adults in an out-patient setting in Maharashtra, India, is an accurate reflection of health in the surrounding community since it was collected at random, with no admission requirements other than age and general good health. Approximately one quarter of all subjects with periodontal disease reported dental problems on the medical health questionnaire, and no group of subjects---even that containing the subjects with the highest MGIs---was more likely to have done so than another (Table 2). More frequent periodontal exams, perhaps of the type described in this study, may keep subjects better apprised of their periodontal and dental health.

Since only 2 subjects had an MGI of 0 and only 1 subject had an MGI of 5, no significant cross-correlations between MGIs of 0 or 5 and any of the questionnaire responses, routine health screening results, or TES results were found (Table 1). A larger sample size would allow more confidence in determining if such correlations are significant or if they are not.

We report for the first time, in an Indian population or otherwise, a significant difference in the periodontal health of males and females: males were more likely to have MGIs of 3 while females were more likely to have MGIs of 2. High prevalence of both gingivitis and more severe periodontal disease in older subjects was found in this study in agreement with previous literature \cite{shewale2016prevalence,shaju2011prevalence}.

\balance

Periodontal health was found to be significantly associated with several diseases and conditions measured in this study. We report significant correlations between an MGI of 4 and swollen joints and a family history of eye diseases. We also found a significant correlation between an MGI of 1 and optic nerve exam abnormalities, reported for the first time in an Indian population (Table 3), which is in agreement with established links between oral and opthalmologic health measured in other populations \cite{pasquale2016prospective,wagley2015periodontal}. We did not find significant correlations between periodontal disease and cardiovascular health. This may be because our screenings only performed a single-lead ECG and otherwise relied on medical history questions as proxies for cardiovascular health. Higher percentages of subjects who reported having diabetes, smoking, and using tobacco were had high MGIs (Table 2) as previously reported by others \cite{loe1993periodontal, shaju2011prevalence}. High and low BMI were both prevalent across all MGIs (Table 3). A relatively low sample size may have contributed to lack of observed cross-correlations in our population. Training a machine learning classifier that can associate image segmentation results with MGI scores and systemic health insights is ongoing in our research group.

\subsection{Summary}

A novel process for oral and TES systemic health screenings and cross-correlations, enabled by imaging, clinical examinations, machine learning, is reported in this manuscript. Association of poor periodontal health with systemic outcomes and poor ophthalmic health reported by us stresses the importance of oral health screenings at the primary care level. Our work shows that one aspect of patient health, such as periodontal health, cannot be fully analyzed in isolation.
The methods and findings communicated in this manuscript can help clinicians and computer scientists in automating the diagnosis and correlation of oral and linked systemic health conditions, ultimately helping patients who might otherwise have limited health care access.

\onecolumn

\section{Appendix}
This appendix provides additional classification results and further granularity for oral-systemic cross-correlations by age and gender.

\setcounter{table}{0}


\begin{table}[!hb]
 \resizebox{\linewidth}{!}{%
 \begin{tabular}{l@{\hskip 0.5cm}cccccccccccccccccccccccccc}
\toprule
MGI & \rotatebox{70}{Glasses} & \rotatebox{70}{Dental}& \rotatebox{70}{Swollen joints} & \rotatebox{70}{Hearing} & \rotatebox{70}{FH diabetes} & \rotatebox{70}{FH high BP} & \rotatebox{70}{Tobacco} & \rotatebox{70}{Difficulty walking} & \rotatebox{70}{High BP} & \rotatebox{70}{Diabetes} & \rotatebox{70}{High BP Rx} & \rotatebox{70}{Asthma} & \rotatebox{70}{Smoking} & \rotatebox{70}{FH cardiac} & \rotatebox{70}{Cardiac Rx} & \rotatebox{70}{Cardiovascular} & \rotatebox{70}{Low BP} & \rotatebox{70}{FH stroke} & \rotatebox{70}{FH eye disease} & \rotatebox{70}{Heart attack} & \rotatebox{70}{Coronary bypass} & \rotatebox{70}{Drinking} & \rotatebox{70}{Eye treatment} & \rotatebox{70}{Memory loss} & \rotatebox{70}{Ear treatment} & \rotatebox{70}{FH ear disease}\\
\midrule
0 & 0  / 1  & 0  / 0  & 0  / 0  & 0 / 0  & 1  / 0  & 0  / 0 & 0 / 1  & 0 / 0  & 0 / 0 & 0 / 1 & 0 / 0 & 0 / 0 & 0 / 0 & 0 / 0 & 0 / 0 & 0 / 0 & 0 / 0 & 0 / 0 & 0 / 0 & 0 / 0 & 0 / 0 & 0 / 0 & 0 / 0 & 0 / 0 & 0 / 0 & 0 / 0\\ 
1 & 15 / 8  & 4  / 7  & 1  / 3  & 2 / 5  & 8  / 5  & 5  / 2 & 0 / 1  & 0 / 3  & 0 / 1 & 0 / 1 & 0 / 1 & 1 / 2 & 0 / 1 & 1 / 2 & 0 / 0 & 0 / 0 & 0 / 0 & 1 / 0 & 0 / 0 & 0 / 0 & 0 / 1 & 0 / 0 & 0 / 0 & 0 / 0 & 0 / 1 & 0 / 0\\ 
2 & 18 / 31 & 10 / 17 & 10 / 13 & 6 / 11 & 11 / 11 & 14 / 8 & 1 / 6  & 5 / 3  & 2 / 4 & 3 / 5 & 2 / 3 & 2 / 3 & 0 / 3 & 0 / 2 & 0 / 2 & 0 / 1 & 1 / 0 & 1 / 1 & 0 / 0 & 0 / 0 & 0 / 0 & 0 / 1 & 0 / 0 & 0 / 0 & 1 / 0 & 0 / 0\\ 
3 & 15 / 28 & 6  / 19 & 9  / 20 & 4 / 18 & 4  / 18 & 4  / 8 & 0 / 11 & 2 / 10 & 3 / 5 & 2 / 5 & 1 / 5 & 0 / 2 & 0 / 2 & 2 / 1 & 0 / 3 & 0 / 1 & 2 / 0 & 1 / 0 & 1 / 0 & 0 / 2 & 0 / 0 & 0 / 1 & 0 / 0 & 0 / 0 & 0 / 0 & 0 / 0\\ 
4 & 6  / 9  & 3  / 5  & \textbf{7$^*$}  / 7  & \textbf{5$^*$} / 6  & 1  / 6  & 0  / 2 & 0 / 3  & \textbf{4$^*$} / 1  & 0 / 2 & 0 / 3 & 1 / 0 & 1 / 0 & 0 / 3 & 0 / 0 & 1 / 0 & 0 / 0 & 0 / 0 & 0 / 0 & 0 / \textbf{2$^*$} & 0 / 0 & 0 / 0 & 0 / 0 & 1 / 0 & 0 / 0 & 0 / 0 & 0 / 1\\ 
5 & 0  / 0  & 0  / 0  & 0  / 0  & 0 / 0  & 0  / 0  & 0  / 0 & 0 / 0  & 0 / 0  & 0 / 0 & 0 / 0 & 0 / 0 & 0 / 0 & 0 / 0 & 0 / 0 & 0 / 0 & 0 / 0 & 0 / 0 & 0 / 0 & 0 / 0 & 0 / 0 & 0 / 0 & 0 / 0 & 0 / 0 & 0 / 0 & 0 / 0 & 0 / 0\\ 
\bottomrule
 \end{tabular}
 }
 \caption{Numbers of subjects of each gender with each modified gingival index (MGI) who responded yes to each question on a medical history questionnaire. Each cell is in the form females / males.
$* p < 0.05$, shown in bold: subjects with the row's MGI and the given gender are more likely to have responded yes to the column's question than subjects of the same gender with other MGIs. BP: blood pressure; FH: family history; Rx: treatment.}
\end{table}

\begin{table}[hb!]
 \resizebox{\linewidth}{!}{%
 \begin{tabular}{l@{\hskip 0.5cm}ccccccccccccc}
\toprule
MGI & \rotatebox{70}{Glasses} & \rotatebox{70}{Dental}& \rotatebox{70}{Swollen joints} & \rotatebox{70}{Hearing} & \rotatebox{70}{FH diabetes} & \rotatebox{70}{FH high BP} & \rotatebox{70}{Tobacco} & \rotatebox{70}{Difficulty walking} & \rotatebox{70}{High BP} & \rotatebox{70}{Diabetes} & \rotatebox{70}{High BP Rx} & \rotatebox{70}{Asthma} & \rotatebox{70}{Smoking}\\
\midrule
0 & 0 / 0 / 1 / 0 & 0 / 0 / 0 / 0 & 0 / 0 / 0 / 0 & 0 / 0 / 0 / 0 & 1 / 0 / 0 / 0 & 0 / 0 / 0 / 0 & 0 / 0 / 1 / 0 & 0 / 0 / 0 / 0 & 0 / 0 / 0 / 0 & 0 / 0 / 1 / 0 & 0 / 0 / 0 / 0 & 0 / 0 / 0 / 0 & 0 / 0 / 0 / 0  \\
1 & 7 / 7 / 8 / 1 & 1 / 3 / 5 / 2 & 0 / 1 / 1 / 2 & 0 / 2 / 3 / 2 & 4 / 7 / 2 / 0 & 3 / 4 / 0 / 0 & 0 / 1 / 0 / 0 & 0 / 0 / 0 / 3 & 0 / 0 / 1 / 0 & 0 / 0 / 1 / 0 & 0 / 0 / 1 / 0 & 0 / 0 / \textbf{3$^*$} / 0 & 0 / 0 / 1 / 0 \\
2 & 10 / 13 / 29 / 7 & 1 / 8 / 14 / 4 & 0 / 5 / 13 / 5 & 1 / 1 / 9 / 6 & 7 / 9 / 13 / 1 & 6 / 8 / 8 / 0 & 0 / 3 / 2 / 2 & 0 / 1 / 4 / 3 & 0 / 0 / 5 / 1 & 0 / 1 / 6 / 1 & 0 / 0 / 4 / 1 & 0 / 1 / 3 / 1 & 0 / 1 / 1 / 1 \\
3 & 3 / 12 / 16 / 12 & 2 / 6 / 12 / 5 & 0 / 1 / 18 / 10 & 0 / 2 / 14 / 6 & 3 / 6 / 3 / 0 & 2 / 6 / 2 / 2 & 0 / 1 / 5 / 5 & 0 / 0 / 9 / 3 & 0 / 0 / 2 / 6 & 0 / 0 / 4 / 3 & 0 / 0 / 1 / 5 & 0 / 0 / 2 / 0 & 0 / 0 / 2 / 0 \\
4 & 0 / 1 / 10 / 4 & 0 / 1 / 5 / 2 & 0 / 1 / 7 / 6 & 0 / 0 / 7 / 4 & 0 / 3 / 2 / 0 & 0 / 1 / 1 / 0 & 0 / 1 / 1 / 1 & 0 / 0 / 4 / 1 & 0 / 0 / 1 / 1 & 0 / 0 / 3 / 1 & 0 / 0 / 1 / 0 & 0 / 0 / 1 / 0 & 0 / 0 / 1 / 2 \\
5 & 0 / 0 / 0 / 0 & 0 / 0 / 0 / 0 & 0 / 0 / 0 / 0 & 0 / 0 / 0 / 0 & 0 / 0 / 0 / 0 & 0 / 0 / 0 / 0 & 0 / 0 / 0 / 0 & 0 / 0 / 0 / 0 & 0 / 0 / 0 / 0 & 0 / 0 / 0 / 0 & 0 / 0 / 0 / 0 & 0 / 0 / 0 / 0 & 0 / 0 / 0 / 0 \\
\bottomrule
 \end{tabular}}
 \bigskip\\
 \resizebox{\linewidth}{!}{%
 \begin{tabular}{l@{\hskip 0.5cm}cccccccccccccccccccccccccc}
\toprule
MGI & \rotatebox{70}{FH cardiac} & \rotatebox{70}{Cardiac Rx} & \rotatebox{70}{Cardiovascular} & \rotatebox{70}{Low BP} & \rotatebox{70}{FH stroke} & \rotatebox{70}{FH eye disease} & \rotatebox{70}{Heart attack} & \rotatebox{70}{Coronary bypass} & \rotatebox{70}{Drinking} & \rotatebox{70}{Eye treatment} & \rotatebox{70}{Memory loss} & \rotatebox{70}{Ear treatment} & \rotatebox{70}{FH ear disease}\\
\midrule
0 & 0 / 0 / 0 / 0 & 0 / 0 / 0 / 0 & 0 / 0 / 0 / 0 & 0 / 0 / 0 / 0 & 0 / 0 / 0 / 0 & 0 / 0 / 0 / 0 & 0 / 0 / 0 / 0 & 0 / 0 / 0 / 0 & 0 / 0 / 0 / 0 & 0 / 0 / 0 / 0 & 0 / 0 / 0 / 0 & 0 / 0 / 0 / 0 & 0 / 0 / 0 / 0 \\
1 &0 / 2 / 1 / 0 & 0 / 0 / 0 / 0 & 0 / 0 / 0 / 0 & 0 / 0 / 0 / 0 & 0 / 1 / 0 / 0 & 0 / 0 / 0 / 0 & 0 / 0 / 0 / 0 & 0 / 0 / 0 / 1 & 0 / 0 / 0 / 0 & 0 / 0 / 0 / 0 & 0 / 0 / 0 / 0 & 0 / 0 / 1 / 0 & 0 / 0 / 0 / 0 \\
2 & 0 / 1 / 1 / 0 & 0 / 0 / 2 / 0 & 0 / 0 / 1 / 0 & 1 / 0 / 0 / 0 & 0 / 1 / 1 / 0 & 0 / 0 / 0 / 0 & 0 / 0 / 0 / 0 & 0 / 0 / 0 / 0 & 0 / 0 / 1 / 0 & 0 / 0 / 0 / 0 & 0 / 0 / 0 / 0 & 0 / 0 / 1 / 0 & 0 / 0 / 0 / 0 \\
3 &0 / 2 / 0 / 1 & 0 / 0 / 0 / 3 & 0 / 0 / 1 / 0 & 0 / 1 / 1 / 0 & 1 / 0 / 0 / 0 & 0 / 0 / 1 / 0 & 0 / 0 / 1 / 1 & 0 / 0 / 0 / 0 & 0 / 0 / 1 / 0 & 0 / 0 / 0 / 0 & 0 / 0 / 0 / 0 & 0 / 0 / 0 / 0 & 0 / 0 / 0 / 0 \\
4 & 0 / 0 / 0 / 0 & 0 / 0 / 1 / 0 & 0 / 0 / 0 / 0 & 0 / 0 / 0 / 0 & 0 / 0 / 0 / 0 & 0 / \textbf{2$^*$} / 0 / 0 & 0 / 0 / 0 / 0 & 0 / 0 / 0 / 0 & 0 / 0 / 0 / 0 & 0 / 0 / 0 / 1 & 0 / 0 / 0 / 0 & 0 / 0 / 0 / 0 & 0 / 1 / 0 / 0 \\
5 & 0 / 0 / 0 / 0 & 0 / 0 / 0 / 0 & 0 / 0 / 0 / 0 & 0 / 0 / 0 / 0 & 0 / 0 / 0 / 0 & 0 / 0 / 0 / 0 & 0 / 0 / 0 / 0 & 0 / 0 / 0 / 0 & 0 / 0 / 0 / 0 & 0 / 0 / 0 / 0 & 0 / 0 / 0 / 0 & 0 / 0 / 0 / 0 & 0 / 0 / 0 / 0 \\
\bottomrule
 \end{tabular}
 }
 \caption{Numbers of subjects of each age cohort with each modified gingival index (MGI) who responded yes to each question on a medical history questionnaire. Each cell is in the form adolescents (18-19) / young adults (20-39) / middle-aged (40-64) / old-aged (65-90). $*p < 0.05$, shown in bold: subjects with the row's MGI and in the given age cohort are more likely to have responded yes to the column's question than subjects in the same age cohort with other MGIs. BP: blood pressure; FH: family history; Rx: treatment.
}
 \end{table}

\begin{table}[ht!]
 \resizebox{\linewidth}{!}{%
 \begin{tabular}{l@{\hskip 1cm}ccccc@{\hskip 1cm}ccccc}
\toprule
& \multicolumn{4}{c}{Routine health screenings} & & \multicolumn{5}{c}{Technology-enabled screenings}\\
MGI & High BP & Low BP & High BMI & Low BMI & & Low O$_2$ & Retinal & TM & Finger-nose & Gait\\
\midrule
0 & 0 / 1 & 0 / 0 & 1 / 1 & 0 / 0 & & 0 / 0 & 0 / 0 & 0 / 0  & 0 / 0 & 0 / 0\\
1 & 1 / 3 & 1 / 0 & 11 / 7 & 3 / 2 & & 1 / 0 & 1 / \textbf{4$^*$} & 2 / 1  & 0 / 0 & 0 / 0\\
2 & 7 / 17 & 0 / 0 & 24 / 31 & 12 / 7 & & 1 / 5 & 0 / 0 & 5 / 3  & 0 / 0 & 0 / 0\\
3 & 1 / 13 & 1 / 1 & 8 / 25 & 6 / 12 & & 1 / 3 & 0 / 0 & 1 / 10 & 0 / 2 & 0 / 1\\
4 & 2 / 7 & 0 / 0 & 4 / 7 & 2 / 5 & & 1 / 0 & 0 / 0 & 1 / 1  & 0 / 0 & 0 / 1\\
5 & 0 / 0 & 0 / 0 & 0 / 0 & 0 / 0 & & 0 / 0 & 0 / 0 & 0 / 0  & 0 / 0 & 0 / 0\\
\bottomrule
 \end{tabular}
 }
 \caption{Numbers of subjects of each gender with each modified gingival index (MGI) who were reported as abnormal in each routine health screening and technology-enabled screening. Each cell is in the form females / males. $*p < 0.05$, shown in bold: subjects with the row's MGI and the given gender are more likely to have the column's condition than subjects of the same gender with other MGIs. BP: blood pressure; BMI: body mass index; O$_2$: blood oxygen level; TM: tympanic membrane.
}
 \end{table}

\newpage

\begin{table}[!ht]
 \resizebox{\linewidth}{!}{%
 \begin{tabular}{l@{\hskip 1cm}ccccc@{\hskip 1cm}ccccc}
\toprule
& \multicolumn{4}{c}{Routine health screenings} & & \multicolumn{5}{c}{Technology-enabled screenings}\\
MGI & High BP & Low BP & High BMI & Low BMI & & Low O$_2$ & Retinal & TM & Finger-nose & Gait\\
\midrule
0 & 0 / 0 / 1 / 0 & 0 / 0 / 0 / 0 & 1 / 0 / 1 / 0 & 0 / 0 / 0 / 0 & & 0 / 0 / 0 / 0 & 0 / 0 / 0 / 0 & 0 / 0 / 0 / 0 & 0 / 0 / 0 / 0 & 0 / 0 / 0 / 0 \\
1 & 0 / 3 / 1 / 0 & 0 / 1 / 0 / 0 & 5 / 6 / 6 / 1 & 4 / 1 / 0 / 0 & & 0 / 0 / 1 / 0 & 1 / 1 / 1 / \textbf{2$^*$} & 0 / 2 / 1 / 0 & 0 / 0 / 0 / 0 & 0 / 0 / 0 / 0 \\
2 & 1 / 6 / 12 / 5 & 0 / 0 / 0 / 0 & 8 / 16 / 28 / 3 & 7 / 9 / 1 / 2 & &  2 / 1 / 2 / 1 & 0 / 0 / 0 / 0 & 1 / 2 / 3 / 2 & 0 / 0 / 0 / 0 & 0 / 0 / 0 / 0 \\
3 & 0 / 1 / 9 / 4 & 1 / 0 / 1 / 0 & 2 / 8 / 16 / 7 & 3 / 4 / 5 / 6 & & 0 / 1 / 1 / 2 & 0 / 0 / 0 / 0 & 2 / 2 / 4 / 3 & 0 / 0 / 2 / 0 & 0 / 0 / 0 / 1 \\
4 & 0 / 1 / 6 / 2 & 0 / 0 / 0 / 0 & 0 / 2 / 8 / 1 & 0 / 2 / 3 / 2 & & 0 / 0 / 1 / 0 & 0 / 0 / 0 / 0 & 0 / 0 / 1 / 1 & 0 / 0 / 0 / 0 & 0 / 1 / 0 / 0 \\
5 & 0 / 0 / 0 / 0 & 0 / 0 / 0 / 0 & 0 / 0 / 0 / 0 & 0 / 0 / 0 / 0 & & 0 / 0 / 0 / 0 & 0 / 0 / 0 / 0 & 0 / 0 / 0 / 0 & 0 / 0 / 0 / 0 & 0 / 0 / 0 / 0 \\
\bottomrule
 \end{tabular}
 }
 \caption{Numbers of subjects of each age cohort with each modified gingival index (MGI) who were reported as abnormal in each routine health screening and technology-enabled screening. Each cell is in the form adolescents (18-19) / young adults (20-39) / middle-aged (40-64) / old-aged (65-90). $* p < 0.05$, shown in bold: subjects with the row's MGI and in the given age cohort are more likely to have the column's condition than subjects in the same age cohort with other MGIs. BP: blood pressure; BMI: body mass index; O$_2$: blood oxygen level; TM: tympanic membrane.}
 \end{table}

\twocolumn
\balance


\newcommand{\BIBdecl}{\setlength{\itemsep}{0.1 em}}
\bibliographystyle{IEEEtran}
\bibliography{sigproc}

\end{document}